\documentclass{INTERSPEECH2023}
\usepackage{multirow}

\interspeechcameraready

\title{Text-only Domain Adaptation using Unified Speech-Text Representation in Transducer}
\name{Lu Huang$^*$, Boyu Li$^*$\thanks{*\ Equal contribution.}, Jun Zhang, Lu Lu, Zejun Ma}
\address{ByteDance}
\email{\{huanglu.thu19,liboyu.622\}@bytedance.com}

\begin{document}
\maketitle
 
\begin{abstract}
Domain adaptation using text-only corpus is challenging in end-to-end(E2E) speech recognition. Adaptation by synthesizing audio from text through TTS is resource-consuming. We present a method to learn Unified Speech-Text Representation in Conformer Transducer(USTR-CT) to enable fast domain adaptation using the text-only corpus. Different from the previous textogram method, an extra text encoder is introduced in our work to learn text representation and is removed during inference, so there is no modification for online deployment. To improve the efficiency of adaptation, single-step and multi-step adaptations are also explored. The experiments on adapting LibriSpeech to SPGISpeech show the proposed method reduces the word error rate(WER) by relatively 44\% on the target domain, which is better than those of TTS method and textogram method. Also, it is shown the proposed method can be combined with internal language model estimation(ILME) to further improve the performance.
\end{abstract}


\section{Introduction}
In recent years, E2E models have achieved significant improvements in automatic speech recognition(ASR)\cite{chiu2018state,sainath2020streaming,li2022recent}. Compared with hybrid models, where acoustic, pronunciation, and language models(LMs) are built and optimized separately, E2E models have achieved promising performance by directly mapping speech features into word sequences. There are some popular E2E models, including connectionist temporal classification\cite{graves2006connectionist,li2018advancing}, recurrent neural network transducer(RNN-T)\cite{sainath2020streaming,li2020developing,graves2012sequence}, and attention-based encoder-decoder\cite{chorowski2015attention,karita2019comparative,li2020comparison}.

However, unlike hybrid models which can adapt to new domains by training LMs using text-only data, E2E model has difficulty in domain adaptation using text-only data. Besides, the E2E models are trained with paired speech-text data, so their generalization ability to different contents is limited, and the performance degrades when a mismatch exists between source and target domains. To overcome this, the most promising approach is to adapt the E2E model using text-only data, because it is much easier to collect text-only data than paired speech-text data in the target domain.

Several methods have been proposed to adapt the E2E model to new domain. The most common solution is to train an external LM using text corpus in the target domain and integrate it into the E2E model during inference, such as shallow fusion\cite{hannun2014deep}, density ratio fusion\cite{mcdermott2019density}, deep fusion\cite{gulcehre2015using}, cold fusion\cite{sriram2017cold}, and internal LM(ILM) estimation based fusion\cite{variani2020hybrid,meng2021internal_ilme}. Nevertheless, all these methods involve an external LM during inference, and the computation cost of decoding is increased. 

Other methods attempt to directly update the E2E model to avoid changing the decoding. Synthesizing paired speech-text data using TTS is a common solution\cite{huang2020rapid,peyser2020improving}, but the process is complex, also the storage and computation cost increases. Recently, a text-to-mel-spectrogram generator is proposed in E2E model\cite{bataev2023text} for replacing TTS. The method 
proposed in \cite{pylkkonen21_interspeech} inserts a temporary LM layer into the prediction network and the LM loss is used for adapting with text-only data. Internal LM adaptation(ILMA)\cite{meng2021internal} is proposed to fine-tune the parameters of internal LM with an additional LM loss. 

Alternative approaches focus on creating a shared embedding space by joint training for two modalities, i.e., speech and text, and have been reported to improve ASR performance without increasing model parameters or decoding complexity\cite{chen2022maestro,sainath2023joist}. Recent works\cite{thomas2022integrating,sato2022text} involve this idea to make a consistent representation between text and speech features in text-only adaptation tasks.

Inspired by \cite{chen2022maestro,thomas2022integrating,sato2022text}, we proposed a method to learn the unified speech-text representation in Conformer Transducer (USTR-CT) for fast text-only domain adaptation. Separated encoders are adopted to learn a shared representation for speech and text features respectively, and a shared encoder is used for the fusion of speech and text representation. At the same time, different representation units are explored and phoneme representation performs best in the target domain. To improve the efficiency of adaptation, single-step and multi-step adaptations are also explored. Finally, We observe 44\% relative WER reduction in the target domain with unpaired text data. In addition, combined with ILME, the proposed method can obtain further gains. 

The paper is organized as follows: Section \ref{related_work} gives a brief introduction about related work. The proposed USTR-CT is discussed in Section \ref{model_architecture}, followed by experiments and discussions in Section \ref{exp_res}.
 
\section{Related work}
\label{related_work}

\subsection{Speech-text joint training for ASR}
Several methods have been proposed to train E2E model with speech and text modalities\cite{chen2022maestro,sainath2023joist,bapna2021slam,bapna2022mslam,tang2022unified,thomas2022towards,chung2021splat,ao2022speecht5}. 

The recent JOIST\cite{sainath2023joist} explores joint training with a combination of losses computed on the supervised paired speech-text data and the unpaired text data in cascaded encoder based streaming ASR framework\cite{narayanan2021cascaded}.
The input unpaired text representation are up-sampled with a simple and parameter-free duration model, and then fed to a text encoder after masking. The output of the text encoder can be fed to the first-pass decoder, or to the shared encoder and second-pass decoder.
However, the experiments were conducted on a multi-domain corpus and the performance of long-tail rare words are evaluated. In this work, we mainly focus on domain adaptation, i.e., transferring the model to the target domain with text-only data.

In order to ensure that the representations learned from speech and text features are aligned, MAESTRO\cite{chen2022maestro} introduces a consistency loss to align two types of representations using the paired data. The method has achieved significant improvements on ASR and speech translation tasks. Different from our work, it focuses on the self-supervised training to obtain a better pre-trained model rather than domain adaptation. Besides, the additional duration model increases the complexity of training.

\subsection{Text-only domain adaptation}

In this subsection, we give a brief introduction to text-only domain adaptation methods without using external LM.

\textbf{TTS adaptation} generates paired speech-text data from the target domain text for fine-tuning\cite{huang2020rapid,peyser2020improving}. However, the number of speakers is limited for TTS and training a reliable multi-speaker TTS model is time-consuming and computationally expensive. Besides, saving the synthesized data also increases storage costs. Moreover, due to the mismatch between synthetic and real audio, additional issues may be also introduced. 
To mitigate the above problems, recently a text-to-spectrogram front-end composed of a text-to-mel-spectrogram generator was introduced in ASR model\cite{bataev2023text}. In this way, on-the-fly generating spectrograms from text-only data is possible, and the mismatch between real and synthetic audio is not obvious as before. However, it still needs to pay attention to the quality of the spectrogram enhancer in the generator during training.

\textbf{ILMA} propose to adapt the E2E model\cite{meng2021internal} by fine-tuning the ILM with text-only data, and parameter regularization is added to avoid over-fitting. Also, it is essential to perform ILM training(ILMT)\cite{meng2021internal_ilmt} in addition to ASR loss before ILMA to ensure that the ILM behaves like a standalone LM.
As only the last linear layer of the jointer network is updated, ILMA's performance on the target domain is also limited.

\textbf{Textogram} was proposed in \cite{thomas2022integrating}, where the text representation is created by repeating one-hot embedding of text tokens by a fixed number of times. The textogram features are stacked together with standard speech features. When training using text-only data, the speech features are set to zero. And when training using paired speech-text data, the textogram features are set to zero.
Due to the concatenation of speech features and textogram, it is necessary to concatenate zero textogram features with speech features during inference. However, in our work, with separated encoders, the input is either text features or speech features during training, and inference can be performed directly without any modifications. Besides, updating only the jointer performs best in textogram, while the jointer, predictor, and even encoder can be adapted to the target domain with better performance. In addition to grapheme representation, subword and phoneme representations are also explored in our work.

\section{Training and adapting methods}
\label{model_architecture}

\subsection{Model architecture}
For the standard RNN-T, speech-text pairs are used to train the model.
Let $\textbf{x}^{\text{speech}}_{1:T}$ be the audio features like Fbank, and $\textbf{y}_{0:u-1}$ be the previous tokens, the output of RNN-T at frame $t$ and step $u$ is computed by
\begin{equation}
    \textbf{h}^{\text{enc}}_{1:T}=\texttt{Encoder}(\textbf{x}^{\text{speech}}_{1:T}),
\end{equation}
\begin{equation}
    \textbf{h}^{\text{pred}}_{u}=\texttt{Predictor}(\textbf{y}_{0:u-1}),
\end{equation}
\begin{equation}
    \textbf{h}^{\text{joint}}_{t,u}=\texttt{Jointer}(\textbf{h}^{\text{enc}}_{t},\textbf{h}^{\text{pred}}_{u}).
\end{equation}

Then with a Softmax layer on the $\textbf{h}^{\text{jointer}}_{t,u}$ and forward-backward algorithm\cite{graves2012sequence}, Transducer loss, which is the sum probability $P(\mathbf{y}|\mathbf{x})$ of all possible alignment paths $\pi$, is computed as the training objective function
\begin{equation}
    \mathcal{L}_{\texttt{rnn-t}}=-\log\sum_{\pi\in\Pi(\mathbf{y})}P(\mathbf{\pi}|\mathbf{x}^{\text{speech}}_{1:T}),
\end{equation}
where $\mathbf{y}=(y_1,\dots, y_U)$ is the label sequence, $U$ is the number of target tokens, and $\Pi(\mathbf{y})$ is the alignment path sets.

\begin{figure}[t]
	\centering
	\includegraphics[width=0.35\textwidth,height=0.35\textwidth]{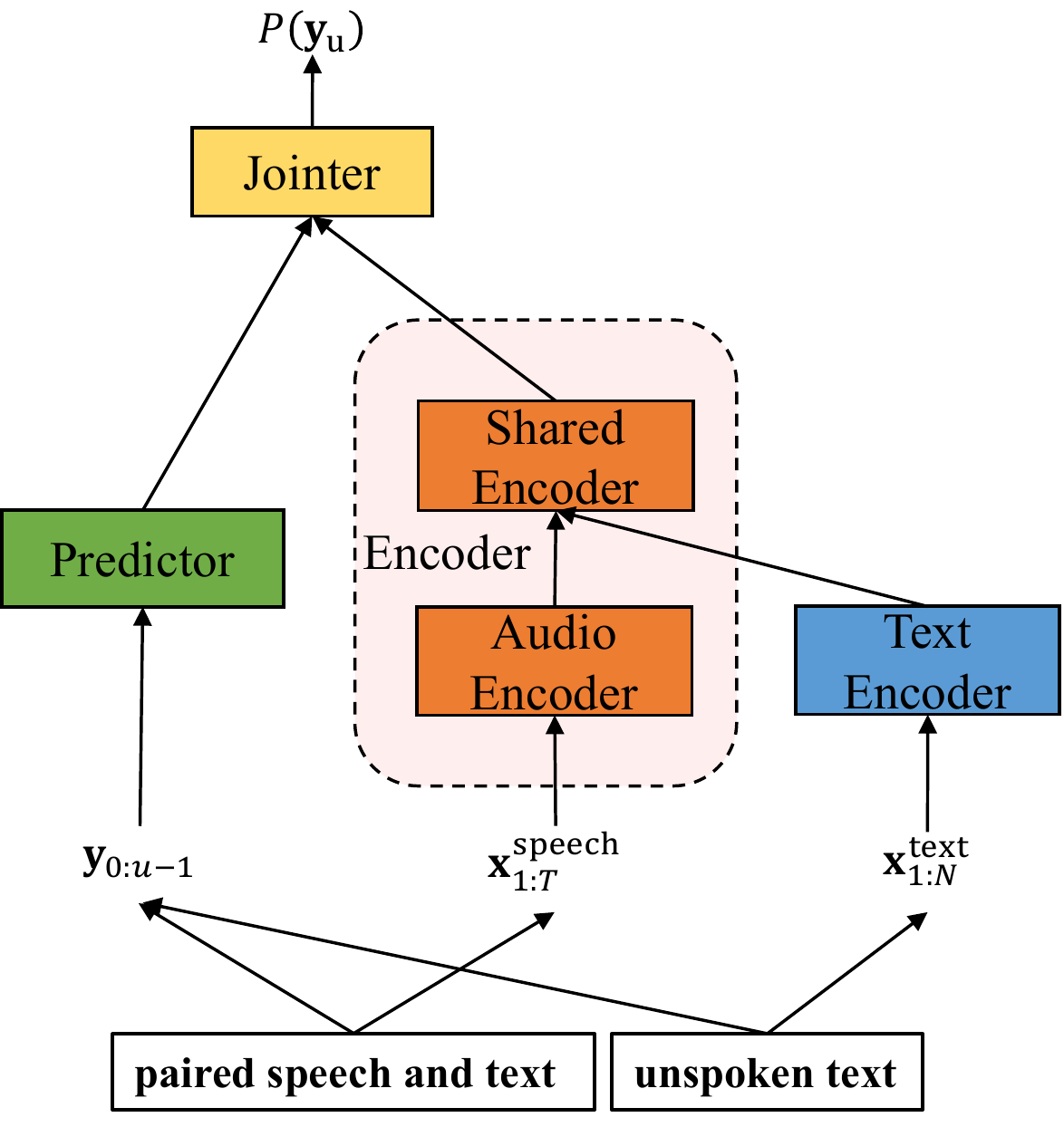}
	\caption{The model structures of USTR-RNN-T.}
	\label{fig:model}
\end{figure}

To involve the text-only corpus during training, the RNN-T encoder is split into two parts, named \texttt{AudioEncoder} and \texttt{SharedEncoder}, and an extra \texttt{TextEncoder} is introduced to model text features $\textbf{x}^{\text{text}}_{1:N}$, which is illustrated in Figure \ref{fig:model} as \textit{USTR-RNN-T}.
The Transducer loss can be computed the same way as paired speech-text corpus,
\begin{equation}
    \textbf{h}^{\text{enc,text}}_{1:N}=\texttt{SharedEncoder}(\texttt{TextEncoder}(\textbf{x}^{\text{text}}_{1:N})),
\end{equation}
\begin{equation}
    \textbf{h}^{\text{joint,text}}_{n,u}=\texttt{Jointer}(\textbf{h}^{\text{enc,text}}_{n},\textbf{h}^{\text{pred}}_{u}),
\end{equation}
where $\textbf{x}^{\text{text}}_{1:N}$ can be grapheme/sub-word/phoneme representations. As the extra \texttt{TextEncoder} can be removed during inference, the proposed method doesn't need any modification for online deployment.

In our experiments, \texttt{SharedEncoder} consists twelve non-streaming Conformer\cite{gulati2020conformer} layers, so the baseline is noted as Conformer Transducer(CT). \texttt{AudioEncoder} has two \texttt{Conv2d} layers with stride of 2 and a linear projection layer, resulting a time reduction of 4. \texttt{TextEncoder} contains an embedding layer and a Transformer layer. For CT’s prediction network, 2-layer LSTM is adopted, and RNN-T’s jointer network is a feed-forward layer.

\subsection{Training}

To enforce speech-text modality matching in a joint embedding space for \texttt{SharedEncoder}, paired speech-text samples are needed for training \texttt{AudioEncoder} and \texttt{TextEncoder}.
When \texttt{TextEncoder} is introduced in the training, the paired speech-text in the training corpus is used as unspoken text with a random probability $p$ by using the text features instead of the audio features.

Three types of text features are considered in this work. The first one is grapheme features, which is similar as the \texttt{textogram} in \cite{thomas2022integrating}. The second one is subword features, which is the same as the output vocabulary of CT, and is generated using \texttt{subword-nmt}\cite{sennrich2016neural}\footnote{https://github.com/rsennrich/subword-nmt}. The final one is phoneme features, which is generated by a Grapheme-to-Phoneme system. For English in this work, \texttt{g2pE}\footnote{https://github.com/Kyubyong/g2p} is used.

To simulate the duration of speech features, the text features are repeated a fixed number of times, which is the same as that in \cite{thomas2022integrating,sainath2023joist}. Also, text features are masked to prevent the \texttt{TextEncoder} from memorizing the grapheme/subword sequence blindly \cite{thomas2022integrating,sainath2023joist}. However, the masking method differs from that in \cite{thomas2022integrating,sainath2023joist} by applying on repeated text features. It is found that masking before repeating brings better performance.

During training, a mini-batch containing both text and speech features is fed into the model. Besides,
ILMT loss is chosen as an optional auxiliary loss, and the overall loss is
\begin{equation}
    \mathcal{L}=\mathcal{L}_{\texttt{rnn-t}} + \lambda\mathcal{L}_{\texttt{ilmt}},
\end{equation}
where $\lambda$ is the weight corresponding to ILMT loss, which is set to 0.2 in all experiments.

\subsection{Adapting}
\begin{figure}[t]
	\centering
	\includegraphics[width=\linewidth]{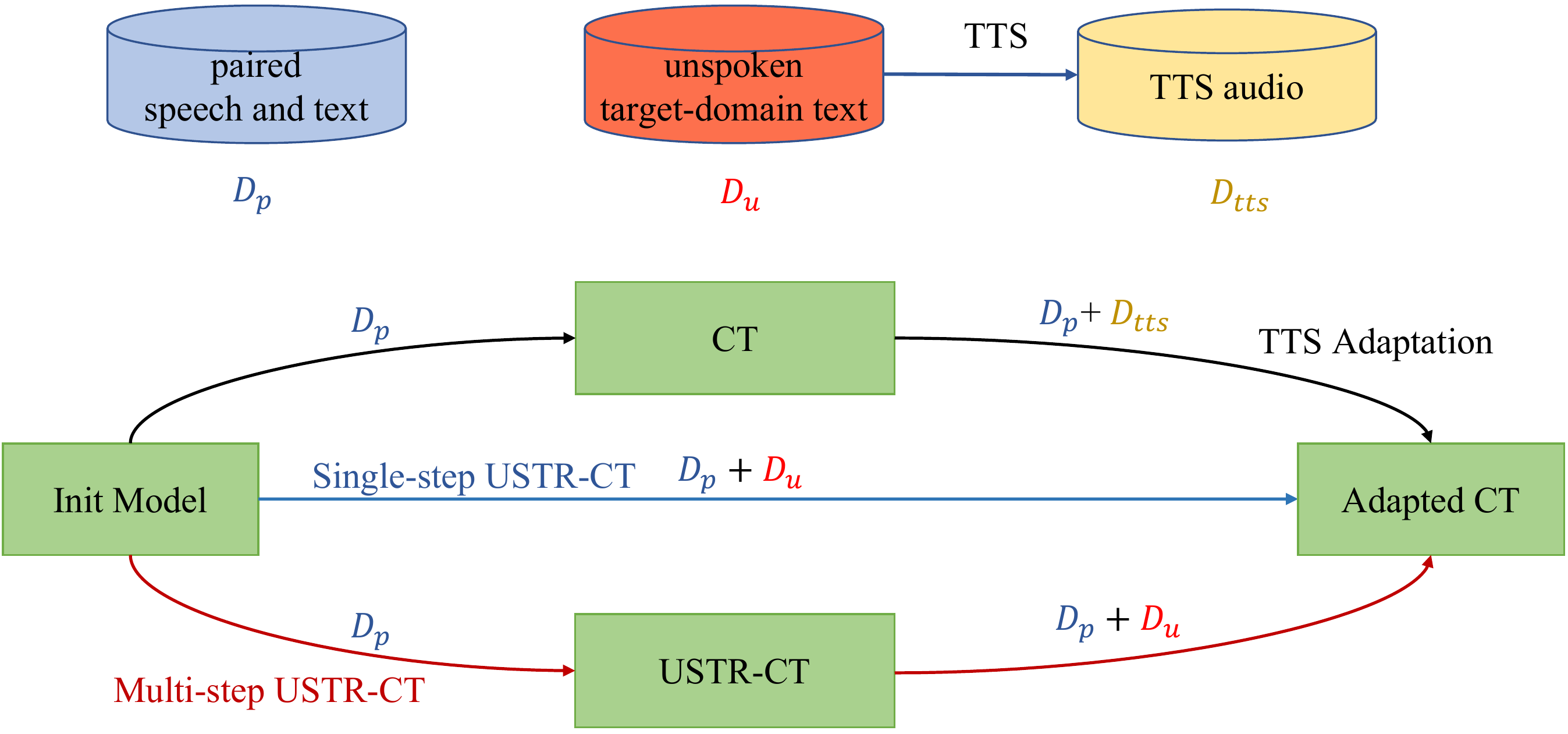}
	\caption{The adaptation processes of TTS, multi-step and single-step USTR-CT.}
	\label{fig:training}
\end{figure}

When the text-only corpus is used for adapting the CT model to a new domain, two adaptation strategies are investigated in this work, as illustrated in Figure \ref{fig:training}. 

\subsubsection{Multi-step adaptation}
As illustrated in the bottom part of Figure \ref{fig:training}, multi-step adaptation using USTR-CT contains two steps.

In the first step, paired speech-text data is used to train a USTR-CT, where each sample is fed into the \texttt{TextEncoder} by using text features instead of audio features with a random probability $p$ to train the \texttt{TextEncoder}. The probability $p$ is set to 0.15 in the experiments.

In the second step, i.e., the stage of adapting, paired speech-text data and unspoken text are both used in each mini-batch with a ratio of 1:1. The ratio can be further tuned to obtain better performance on the target domain, which is left for future discussion. The paired speech-text data is used to maintain the performance on source domain. In this step, the parameters of \texttt{AudioEncoder} and \texttt{SharedEncoder} (i.e., the \texttt{Encoder} of CT) are kept constant, while \texttt{Jointer} and \texttt{Predictor} are trained to adapt to new domain. Due to the existence of USTR-CT model after first step, it is more convenient for adapting to other domains when there is multi-domain scenario.

\subsubsection{Single-step adaptation}
As shown in the middle part of Figure \ref{fig:training}, single-step USTR-CT trains an adapted CT model from random initialization directly. Similar to multi-step USTR-CT, paired speech-text data is also fed into the \texttt{TextEncoder} by a probability $p=0.15$. Also, the ratio between paired speech-text data and unspoken text is still 1:1 to be consistent with multi-step adaptation.

\section{Experiments and results}
\label{exp_res}
\subsection{Experimental setup}
The experiments are conducted on LibriSpeech\cite{panayotov2015librispeech} and SPGISpeech\cite{o2021spgispeech} corpora.
SPGISpeech contains 5,000 hours of financial audio. In this work, only the transcribed text of SPGISpeech is used for text-only domain adaptation, which has 1.7M utterances. Two versions of the text are created in the experiments, noted as \texttt{Large}(L) and \texttt{Small}(S), where the former contains the full 1.7M utterances and the latter contains a subset of 280k utterances, which is almost the same as the number of Librispeech utterances. Besides, the TTS audios are synthesized from the \texttt{Small} subset using an in-house engine to indicate that TTS is resource-consuming.

For the audio features, the 80-dim filter-bank(Fbank) is used and Spec-Augment\cite{park2019specaugment} is applied on Fbank features. 
The text features are masked with a probability of 0.15 before repeating.
The model's structure is described as that in Section \ref{model_architecture}, and the output of RNN-T is 4,048 subword units. All models are trained with PyTorch\cite{paszke2019pytorch}. WER is evaluated on Librispeech \texttt{test-clean}/\texttt{test-other} sets and SPGISpeech \texttt{val} set to measure the ASR performance on source and target domain. 

\subsection{Baseline systems}
A CT model is trained on LibriSpeech, which achieves a WER of 23.55\% on SPGISpeech \texttt{val} set. With TTS based adaptation, as shown in the top part of Figure \ref{fig:training}, the WER is reduced to 14.99\% by 36.35\% relatively. Besides, the textogram\cite{thomas2022integrating} method is also evaluated in this work, which achieves a WER of 23.94\%. Textogram based adaptation, where the encoder and jointer are kept constant, is trained with text-only corpus. And after adaptation, the WER is reduced by relatively 33.25\%/37.80\% when using the S/L subset respectively.

\begin{table}[th]
  \caption{The WER(\%) of different systems on LibriSpeech test sets and SPGISpeech val set. Text adaptation L/S corresponds to the \texttt{Large}/\texttt{Small} subset of SPGISpeech's transcribed text.}
  \label{tab:baseline}
  \centering
  \begin{tabular}{l|c|c}
    \toprule
    \multirow{2}{*}{model} & LibriSpeech test &  SPGISpeech \\
     & clean/other & val \\
    \midrule
    CT  &     3.99/8.28 & 23.55      \\
    \midrule
    \ \ + TTS adaptation & 3.85/8.12 &14.99 \\
    \midrule
    textogram baseline \cite{thomas2022integrating} &4.18/8.84 & 23.94  \\
    \ \ + text adaptation(S) &  5.12/10.10   &  15.98   \\
    \ \ + text adaptation(L) &  4.43/8.98  &  14.89   \\
    \midrule
    multi-step USTR-CT & 3.76/8.15   &  22.72  \\
    \ \ + text adaptation(S) &   3.66/8.00   &  14.89   \\
    \ \ + text adaptation(L) &   \textbf{3.64/7.84}   &  \textbf{14.61}   \\
    \bottomrule
  \end{tabular}
\end{table}

The proposed multi-step USTR-CT is firstly trained with grapheme representation by masking with a rate of 0.15 and repeating four times. As illustrated in Table \ref{tab:baseline}, the proposed USTR-CT achieves a WER of 22.72\% on SPGISpeech \texttt{val} set before adaptation, which is better than the textogram. After adaptation, the proposed method not only performs better on the target domain, but also achieves the best performance on LibriSpeech test sets, as the paired speech-text data is used to maintain the performance of source domain during adaptation. The results indicate that the extra text encoder in USTR-CT and jointer adaptation in the second step are beneficial. Also, the proposed method outperforms TTS based adaptation on both source and target domains, with relative WER reductions of 2.54\%$\sim$5.45\%.

\subsection{Representation units}
\begin{table}[th]
  \caption{The WER(\%) of multi-step USTR-CT using different representation units for text features.}
  \label{tab:rep_char}
  \centering
  \begin{tabular}{l|c|c}
    \toprule
    \multirow{2}{*}{model} & LibriSpeech test &  SPGISpeech \\
     & clean/other & val \\
    \midrule
    grapheme repeat 4 & 3.76/8.15   &  22.72  \\
    \ \ + text adaptation(L) &   \textbf{3.64}/7.84   &  14.61   \\
    \midrule
    phoneme repeat 3 & 4.10/8.32 & 23.11  \\
    \ \ + text adaptation(L) &   3.80/7.96   & 13.41   \\
    \midrule
    phoneme repeat 4 &3.82/8.18 & 22.30  \\
    \ \ + text adaptation(L) & \textbf{3.64/7.80} & \textbf{13.38}   \\
    \midrule
    phoneme repeat 5 & 3.82/8.08 & 22.17  \\
    \ \ + text adaptation(L) &  3.71/7.82   &  13.42   \\
     \midrule
    subword repeat 4 & 4.20/8.49 & 23.92  \\
    \ \ + text adaptation(L) &   3.80/8.27   &  15.00   \\
    \bottomrule
  \end{tabular}
\end{table}

Different representation units are explored for multi-step USTR-CT, where the repeating number of phoneme representation is also investigated. As illustrated in Table \ref{tab:rep_char}, for the repeating number of 4, phoneme representation performs best on target domain(WER 13.38\% vs. 14.61\%/15.00\%). This may due to the phoneme representation is more relevant to Fbank features, and thus the learning of unified speech-text representation can be more easier. Besides, we changed the repeating number from 4 to 3/5 and no further gains were observed.

\subsection{Multi-step vs. single-step}
\label{multi_vs_single}
We compared single-step USTR-CT with multi-step USTR-CT and TTS based adaptation, and the results are illustrated in Table \ref{tab:single_step}. It is shown that single-step USTR-CT performs best on both source and target domain. As the shared encoder is also adapted to target domain, the single-step USTR-CT performs even better than multi-step USTR-CT. Besides, the extra text of SPGISpeech also benefits the source domain.

\begin{table}[th]
  \caption{The WER(\%) of multi-/single-step USTR-CT.}
  \label{tab:single_step}
  \centering
  \begin{tabular}{l|c|c}
    \toprule
    \multirow{2}{*}{model} & LibriSpeech test &  SPGISpeech \\
     & clean/other & val \\
    \midrule
    CT + TTS adaptation &     3.85/8.12 & 14.99      \\
    \midrule
    multi-step USTR-CT & 3.82/8.18 & 22.30  \\
    \ \ + text adaptation(L) &   3.64/7.80   &  13.38   \\
    \midrule
    single-step USTR-CT(L)  &   \textbf{3.07/7.13}   &  \textbf{13.25}   \\
    \bottomrule
  \end{tabular}
\end{table}

\subsection{Combination with ILME}
\begin{table}[th]
  \caption{The WER(\%) of different models with ILME.}
  \label{tab:ilme}
  \centering
  \begin{tabular}{l|c}
    \toprule
    model &  SPGISpeech val \\
    \midrule
    CT & 23.55      \\
    \ \ + ILME &  13.83   \\
    \midrule
    CT + TTS adaptation & 14.99      \\
    \ \ + ILME &  11.34   \\
    \midrule
    multi-step USTR-CT &  13.38   \\
    \ \ + ILME   &  \textbf{10.05}   \\
    \midrule
    single-step USTR-CT    &  13.25   \\
    \ \ + ILME    &  10.80   \\
    \bottomrule
  \end{tabular}
\end{table}

As the text of target domain is already involved in training of the CT model, the benefit of external LM may be discounted. We have trained an LSTM LM and evaluated the performance of ILME using different CT models. As illustrated in Table \ref{tab:ilme}, ILME brings WER reductions of 41.27\%/24.35\% on baseline CT and TTS based models. For multi-/single-step USTR-CT, ILME also reduces the WER by 24.89\%/18.49\% respectively. This indicated that the proposed USTR is able to combine with ILME to further improve the performance on the target domain.

It is noticed that multi-step USTR-CT performs better than single-step USTR-CT when combined with ILME, which is different from the results without ILME in Section \ref{multi_vs_single}. We assume that the ILM score of single-step USTR-CT is not accurate, as the encoder also captures the linguistic information of target domain during training. Besides, as the encoder is frozen during the 2nd step, multi-step USTR-CT is more suitable for training with a large scale text corpus.

\section{Conclusions}
In this work, an extra text encoder is introduced for text-only domain adaptation, which outperforms the TTS adaptation by 11.61\% relatively when using phoneme representation. Compared to TTS adaptation, the proposed USTR-CT is efficient and resource-saving for fast domain adaptation. Besides, USTR-CT is able to adapt to the target domain with a single-step training and combine with ILME to obtain further gains. Although experiments were conducted on non-streaming models in this work, the method is still applicable for streaming ASR. With the separated speech and text encoders, the similarity between the speech and text modalities can be considered to further improve the performance, which is left as a discussion in future work.

%
%
%
%

\bibliographystyle{IEEEtran}
\bibliography{mybib}

\end{document}